\documentclass[lettersize,journal]{IEEEtran}
\usepackage[table]{xcolor}
\usepackage[utf8]{inputenc}
\usepackage{amsmath,amsfonts}
\usepackage{algorithmic}
\usepackage{algorithm}
\usepackage{array}
\usepackage{subfigure}
\usepackage{textcomp}
\usepackage{stfloats}
\usepackage{url}
\usepackage{verbatim}
\usepackage{graphicx}
\usepackage{cite}
\usepackage{color}
\usepackage{booktabs}
\usepackage{tabularx}
\usepackage{multirow}
\usepackage{utfsym}
\usepackage{hyperref}
\usepackage{makecell}

\definecolor{barrier}{RGB}{242,122,52}
\definecolor{bicycle}{RGB}{237,186,191}
\definecolor{bus}{RGB}{255,255,79}
\definecolor{car}{RGB}{0,134,207}
\definecolor{constveh}{RGB}{0,222,222}
\definecolor{motorcycle}{RGB}{161,141,0}
\definecolor{pedestrian}{RGB}{230,32,20}
\definecolor{trafficcone}{RGB}{218,207,125}
\definecolor{trailer}{RGB}{109,58,0}
\definecolor{truck}{RGB}{160,89,255}
\definecolor{drivesurf}{RGB}{255,0,255}
\definecolor{otherflat}{RGB}{122,122,122}
\definecolor{sidewalk}{RGB}{46,10,71}
\definecolor{terrain}{RGB}{158,239,112}
\definecolor{manmade}{RGB}{201,201,201}
\definecolor{vegetation}{RGB}{0,158,0}

\begin{document}

\title{StereoMV2D: A Sparse Temporal Stereo-Enhanced Framework for Robust Multi-View 3D Object Detection}

\author{Di Wu, Feng Yang, \textit{Member, IEEE,}Wenhui Zhao, Jinwen Yu, Pan Liao, Benlian Xu, Dingwen Zhang, \textit{Member, IEEE,}

\thanks{This work was supported in part by the National Natural Science
Foundation of China (No. 62576238, 62293543, U24A20263) and the Suzhou municipal science and technology plan project (No. SYG202351). (Corresponding author: Feng Yang and Benlian Xu.)}

\thanks{Di Wu, Feng Yang, Wenhui Zhao, Jinwen Yu,  Pan Liao and Dingwen Zhang are with the school of automation, Northwestern Polytechnical University, Xi'an Shanxi 710072, China (e-mail: wu\_di821@mail.nwpu.edu.cn; yangfeng@nwpu.edu.cn; zwh2024202513@mail.nwpu.edu.cn; yujinwen@mail.nwpu.edu.cn; liaopan@mail.nwpu.edu.cn;  zdw2006yyy@nwpu.edu.cn.}
\thanks{Benlian Xu is with the school of electronic and information engineering, Suzhou University of Science and Technology, Suzhou Jiangsu 215009, China (e-mail:xu\_benlian@usts.edu.cn).}
        }

\markboth{Manuscript}%
{Shell \MakeLowercase{\textit{et al.}}: A Sample Article Using IEEEtran.cls for IEEE Journals}

\maketitle

\begin{abstract}
Multi-view 3D object detection is a fundamental task in autonomous driving perception, where achieving a balance between detection accuracy and computational efficiency remains crucial. Sparse query-based 3D detectors efficiently aggregate object-relevant features from multi-view images through a set of learnable queries, offering a concise and end-to-end detection paradigm. Building on this foundation, MV2D leverages 2D detection results to provide high-quality object priors for query initialization, enabling higher precision and recall. However, the inherent depth ambiguity in single-frame 2D detections still limits the accuracy of 3D query generation. To address this issue, we propose StereoMV2D, a unified framework that integrates temporal stereo modeling into the 2D detection–guided multi-view 3D detector. By exploiting cross-temporal disparities of the same object across adjacent frames, StereoMV2D enhances depth perception and refines the query priors, while performing all computations efficiently within 2D regions of interest (RoIs). Furthermore, a dynamic confidence gating mechanism adaptively evaluates the reliability of temporal stereo cues through learning statistical patterns derived from the inter-frame matching matrix together with appearance consistency, ensuring robust detection under object appearance and occlusion. Extensive experiments on the nuScenes and Argoverse 2 datasets demonstrate that StereoMV2D achieves superior detection performance without incurring significant computational overhead. Code will be available at https://github.com/Uddd821/StereoMV2D.
\end{abstract}


\section{Introduction}
\IEEEPARstart{M}ulti-view 3D object detection constitutes one of the fundamental and core tasks underlying numerous frontier technologies operating in the three-dimensional world, such as autonomous driving and robotic navigation\cite{ma2024vision, li2025vdg, zhou2025state}. These real-world application systems inherently impose strict requirements on real-time performance and safety, thereby demanding both high accuracy and computational efficiency from perception models\cite{chang2024unified}. Consequently, a major research challenge arises: how to achieve precise and reliable 3D object detection while maintaining high inference efficiency, ensuring that perception systems can operate seamlessly in dynamic and safety-critical environments. Addressing this trade-off between accuracy and efficiency has thus become a central issue in the advancement of modern 3D perception frameworks\cite{li2024fast, wang2023exploring, liu2023sparsebev}.

Transformer-based query methods have gradually become dominant in recent 3D object detection due to their concise and efficient end-to-end paradigm\cite{li2024bevformer, yang2024widthformer, wu2025hv, wang2022detr3d, chu2025oa}. Among them, sparse query–based approaches predefine a set of learnable queries distributed in 3D space, which interact with image features through 3D–2D coordinate transformations to iteratively update the query features, thereby directly learning effective 3D object representations, as illustrated in Fig. 1(a). Originating from the DETR family of detectors\cite{carion2020end, zhu2020deformable, zhang2022dino, zhao2024detrs}, these methods naturally inherit similar limitations: the complexity introduced by high-resolution feature maps makes it difficult for queries to focus on sparse and semantically meaningful regions, leading to low training efficiency and slow convergence\cite{zhu2020deformable}. To address these issues, numerous improvements have been proposed in the 2D detection domain, which can be broadly categorized into two directions: (1) Token sparsification, which reduces computational cost by filtering out redundant regions in feature maps\cite{roh2021sparse}; and (2) Query selection, which provides positional priors for initializing object queries, effectively mitigating convergence bottlenecks during training and guiding feature localization\cite{liu2022dab, zhang2022dino, zhao2024detrs}. 

\begin{figure*}[!t]
\centering
\subfigure[DETR3D]{
	\includegraphics[height=0.139\textheight]{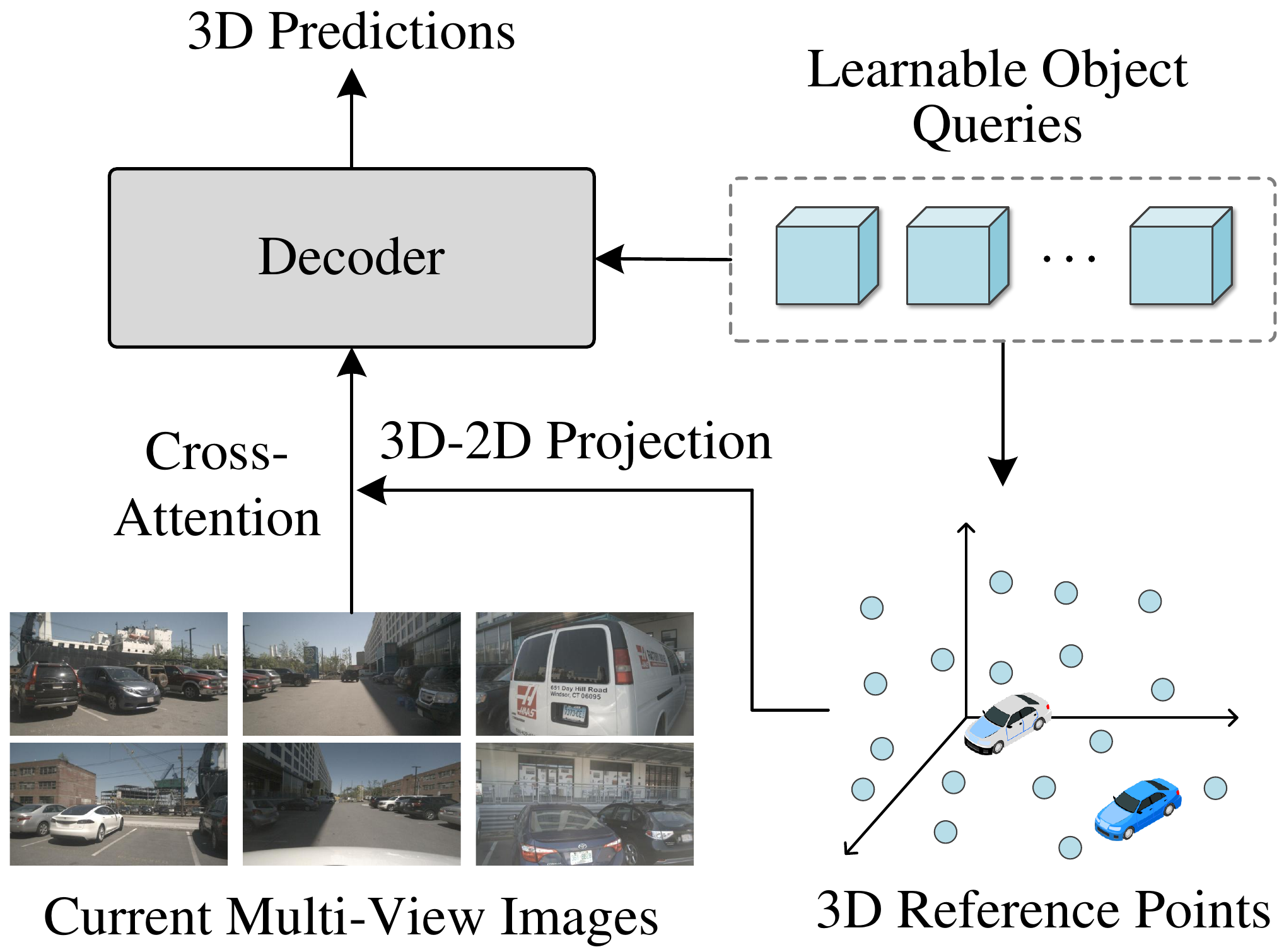}
	\label{fig:a}}
\subfigure[MV2D]{
	\includegraphics[height=0.139\textheight]{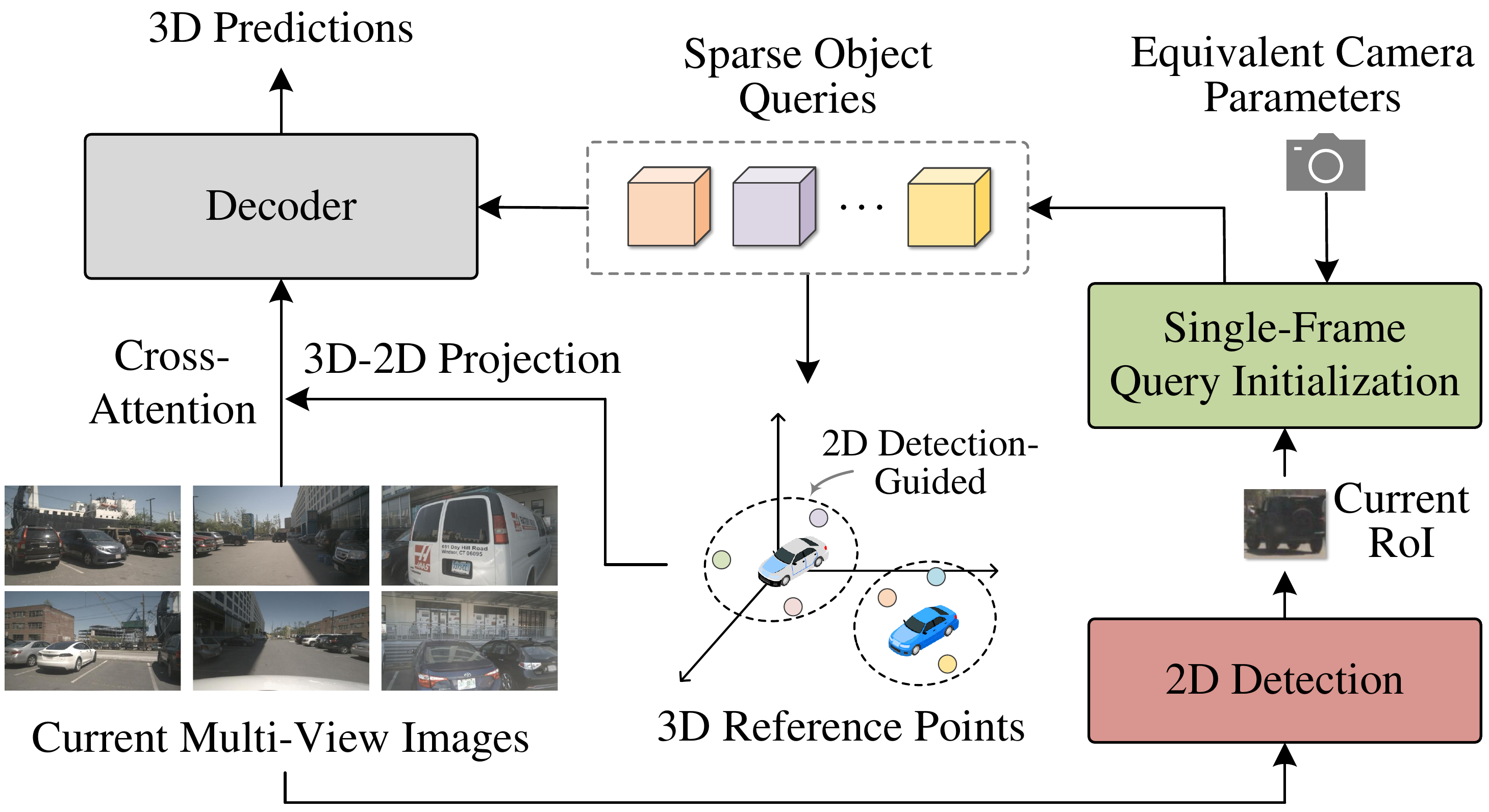}
	\label{fig:b}}
\subfigure[StereoMV2D]{
	\includegraphics[height=0.139\textheight]{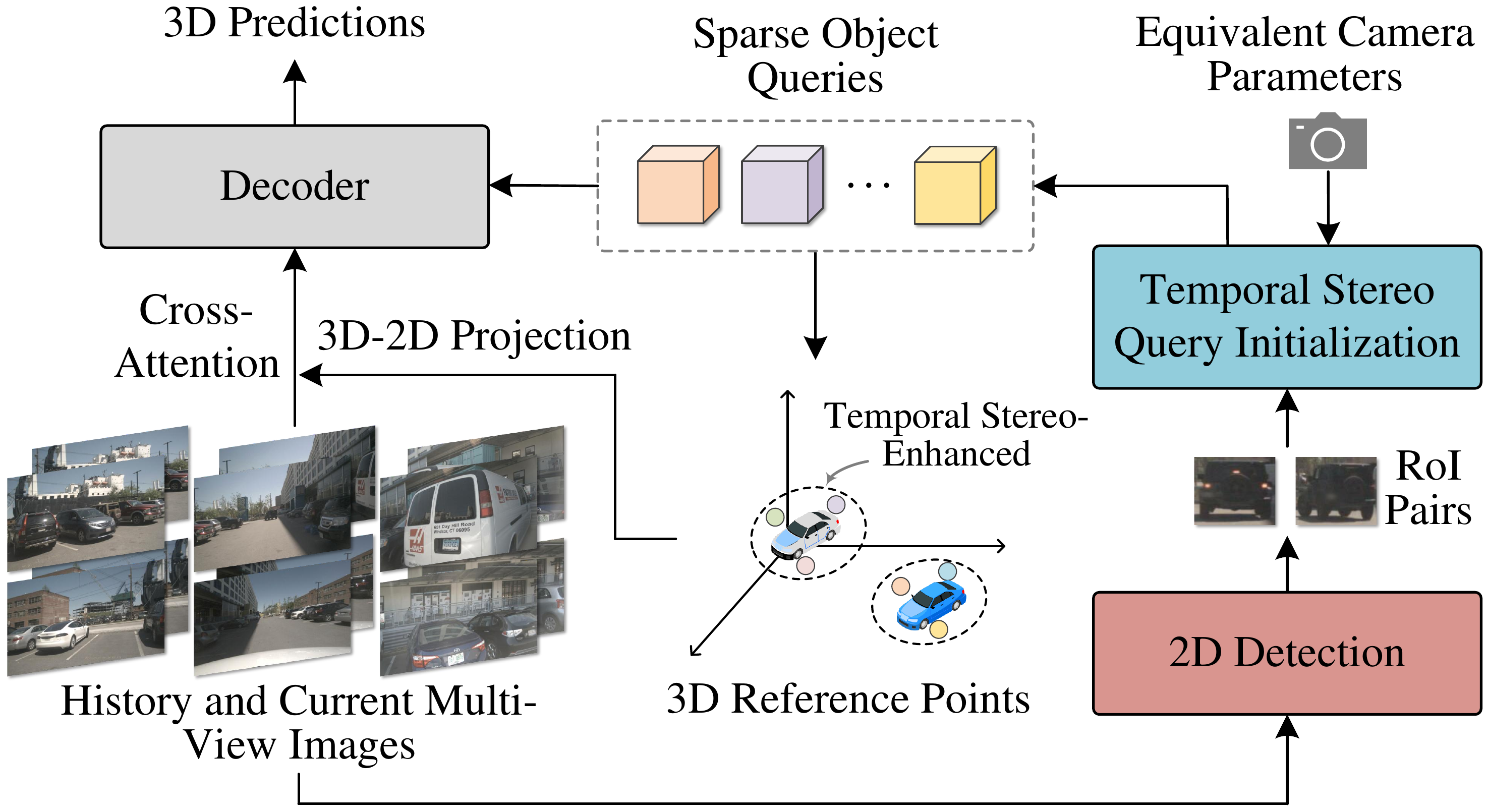}
	\label{fig:c}}
\caption{Comparison of different paradigms for query-based multi-view 3D object detection. (a) DETR3D adopts numerous learnable queries without any location priors. (b) MV2D initializes sparse queries using 2D detections, improving localization but still suffering from monocular depth ambiguity. (c) StereoMV2D (ours) integrates RoI-level temporal stereo into a sparse query framework to enhance depth reasoning. }
\label{fig_1}
\end{figure*}

Following these two main directions, several sparse query-based 3D detection frameworks have emerged to improve the representation capability and efficiency of query learning. For example, ToC3D \cite{zhang2024make} adopts the idea of token sparsification, dynamically filtering redundant regions from multi-view feature maps to reduce computational overhead while preserving discriminative spatial information. In contrast, QAF2D \cite{ji2024enhancing} and MV2D \cite{wang2023object} fall into the category of query selection–based methods, which leverage high-quality 2D detections to provide strong positional priors for initializing 3D queries, as depicted in Fig. 1(b). This strategy enables more accurate object localization and higher recall, particularly for small and distant objects, by ensuring that the generated queries effectively cover most instances in the scene. Among them, MV2D achieves a particularly effective integration of 2D detection guidance and multi-view feature aggregation. By introducing 2D object detectors to initialize 3D queries, it bridges the semantic gap between 2D perception and 3D reasoning, and thus enhances the overall training efficiency. However, its dependence on single-frame 2D detections inevitably suffers from depth ambiguity—errors in monocular depth estimation can propagate into 3D query initialization, leading to inaccurate geometric priors and degraded feature interaction. Several recent studies have recognized this problem and incorporated temporal stereo modeling into multi-view 3D object detection. BEVStereo\cite{li2023bevstereo}, SOLOFusion\cite{park2022time}, and STS\cite{wang2022sts} introduce cross-frame geometric constraints to construct cost volumes for predicting depth maps, which are then used to refine BEV features. In addition, CVT-Occ\cite{ye2024cvt} completes ambiguous depth representations by projecting discretely sampled 3D points along camera rays onto images from different timestamps, thereby enhancing temporal depth consistency. Although these methods effectively improve depth estimation accuracy, they typically rely on constructing dense cost volumes over the entire image domain, which introduces substantial computational and memory overhead. Meanwhile, sparse query–based approaches inherently emphasize efficient, object-level feature aggregation. This naturally raises an important question: Can temporal stereo geometry be incorporated to alleviate single-frame depth ambiguity while preserving the sparse and efficient nature of these methods? 

Motivated by this insight, we propose StereoMV2D, a novel temporal stereo–driven sparse query framework that enhances depth perception in multi-view 3D detection while preserving the efficiency advantages of sparse query–based methods, as illustrated in Fig. 1(c). Building upon the MV2D paradigm, StereoMV2D embeds temporal geometric consistency into query initialization to improve 3D localization accuracy without introducing significant computational cost. Specifically, it leverages cross-temporal disparities of the same object across adjacent frames to alleviate the inherent ambiguity in monocular depth estimation, thereby generating more accurate and stable depth priors for query initialization. Unlike traditional dense stereo matching, our method generates RoI-level stereo cost volumes within 2D RoIs, which restricts temporal correspondence to local object areas, significantly reducing computation while preserving geometry-aware stereo cues. Furthermore, to ensure robustness under dynamically changing scenes—where objects may appear, disappear, or undergo occlusion—StereoMV2D employs a lightweight confidence gating mechanism that estimates the reliability of temporal stereo cues from soft matching statistics and appearance consistency, and adaptively fuses them into the query refinement. Through this unified design, StereoMV2D achieves a balanced integration of depth-aware accuracy, temporal robustness, and inference efficiency, establishing an effective and scalable framework for real-world multi-view 3D object detection.

Our contributions can be concluded as follows:
\begin{itemize}
\item We propose StereoMV2D, a novel 3D object detection framework that integrates RoI-level temporal stereo with sparse query–based detection, enhancing depth reasoning without introducing expensive computation cost. \item To handle the dynamic and uncertain nature of real-world scenes, we introduces a confidence learning strategy that evaluates the reliability of temporal stereo cues through inter-frame matching statistics and feature consistency. \item Extensive experiments conducted on the competitive nuScenes\cite{caesar2020nuscenes} and Argoverse 2\cite{wilson2023argoverse} datasets demonstrate the effectiveness of the proposed method, achieving a favorable balance between accuracy and efficiency.
\end{itemize}

\section{Related Work}
\subsection{2D Object Detection}
2D object detection has long been a fundamental problem in the development of computer vision. Early approaches were predominantly based on convolutional neural networks (CNNs) and can be broadly categorized into two-stage\cite{ren2016faster, he2017mask} and one-stage\cite{tian2019fcos, zhou2019objects, lin2017focal} detection paradigms, depending on whether they first generate region proposals that potentially contain objects. With the remarkable success of Transformers in natural language processing (NLP)\cite{vaswani2017attention}, DETR\cite{carion2020end} pioneered the introduction of the Transformer encoder–decoder architecture into object detection and achieved impressive performance. However, as a fully end-to-end paradigm, DETR suffers from slow convergence and limited feature localization ability, mainly due to the difficulty of learning global attention from dense feature maps without strong spatial priors. To address these limitations, numerous variants have been proposed to enhance its training efficiency and localization precision. Deformable DETR \cite{zhu2020deformable} adopts multi-scale deformable attention to focus on sparse, informative regions. Conditional DETR \cite{meng2021conditional}, Anchor DETR \cite{wang2022anchor}, and DAB-DETR \cite{liu2022dab} enhance query position encoding using anchor- or box-based priors. DN-DETR \cite{li2022dn} introduces denoising training to stabilize bipartite matching, and DINO \cite{zhang2022dino} further boosts convergence with multi-scale denoising queries. RT-DETR\cite{zhao2024detrs} redesigns the encoder–decoder interaction for real-time detection. These advancements have made transformer-based detectors increasingly competitive with conventional CNN-based approaches, laying a solid foundation for extending query-based paradigms into more complex perception tasks such as 3D object detection.

\subsection{Multi-view 3D Object Detection}
For safety-critical applications such as autonomous driving, modern perception systems are typically equipped with multiple cameras covering complementary viewpoints, which has driven increasing interest in multi-view 3D object detection. Early camera-based methods, such as Lift-Splat-Shoot \cite{philion2020lift}, lift image features into 3D space using estimated depth distributions and project them into a unified coordinate frame via camera geometry. Subsequent works \cite{huang2021bevdet, li2023bevdepth, chi2023bev} have standardized this pipeline into a four-stage BEV-based architecture consisting of an image encoder, a view transformer, a BEV encoder, and a detection head. To alleviate error accumulation caused by ill-posed depth estimation, Transformer-based detectors \cite{li2024bevformer, wang2022detr3d} replace explicit depth lifting with attention-driven feature aggregation. These approaches can be broadly categorized into dense BEV query–based methods \cite{li2024bevformer, yang2024widthformer, pan2024clip}, which refine a fixed BEV grid via cross-attention, and sparse object query–based methods \cite{wang2023exploring, wang2022detr3d, lin2022sparse4d}, which directly model foreground instances using a limited number of learnable 3D queries. The latter paradigm has demonstrated superior efficiency and competitive accuracy by focusing computation on object-centric regions \cite{lin2022sparse4d, liu2023sparsebev, wang2023exploring}. Recent studies have further explored temporal modeling to enhance depth perception in multi-view 3D detection. BEV-based frameworks such as BEVStereo \cite{li2023bevstereo} and STS \cite{wang2022sts} explicitly construct cross-frame cost volumes to enforce temporal stereo consistency. While effective, these methods rely on dense BEV representations and incur substantial computational overhead. In parallel, sparse query–based detectors have introduced temporal object propagation mechanisms \cite{wang2023exploring, lin2022sparse4d} to maintain object continuity across frames, yet such designs primarily emphasize appearance-level consistency rather than explicit geometric stereo reasoning. To the best of our knowledge, no prior work explicitly integrates temporal stereo modeling into sparse query–based 3D detection, achieving cross-frame depth reasoning while preserving object-level efficiency. This gap motivates our proposed StereoMV2D. 

\subsection{Multi-View Stereo Matching}
Multi-view stereo (MVS) aims to reconstruct scene geometry from multiple calibrated images by estimating dense depth maps with cross-view photometric consistency. Traditional MVS pipelines relied on hand-crafted similarity metrics and geometric optimization \cite{furukawa2009accurate, schonberger2016pixelwise}, but suffered from sensitivity to textureless regions and occlusions. With the success of deep learning, MVSNet \cite{yao2018mvsnet} pioneered the use of a differentiable cost volume for learning multi-view depth estimation in an end-to-end manner. Following MVSNet, numerous variants have been proposed to enhance performance and reduce computational overhead. R-MVSNet \cite{yao2019recurrent} replaces the full 3D convolution with recurrent GRU-based regularization to lower memory usage. CasMVSNet \cite{gu2020cascade} adopts a coarse-to-fine cascade structure that progressively refines depth estimation at multiple resolutions, achieving a better balance between accuracy and efficiency. Fast-MVSNet \cite{yu2020fast} further optimizes the network architecture and sampling strategy for real-time applications. MVSTER \cite{wang2022mvster} introduces an epipolar transformer to model multi-view feature correlations along geometric constraints, achieving more efficient and accurate depth estimation without relying on dense 3D convolutions. These advances mark a shift from volumetric and memory-intensive designs toward lightweight, transformer-based MVS frameworks. In parallel, the idea of temporal stereo has also been successfully adopted in 3D object detection, as demonstrated by BEVStereo \cite{li2023bevstereo} and SOLOFusion \cite{park2022time}, which perform multi-view temporal stereo to improve depth estimation. However, these approaches rely on dense BEV feature representations, leading to substantial computational cost.

\section{Methods}

\subsection{Overview}
\begin{figure*}[!t]
\centering
\includegraphics[width=7in]{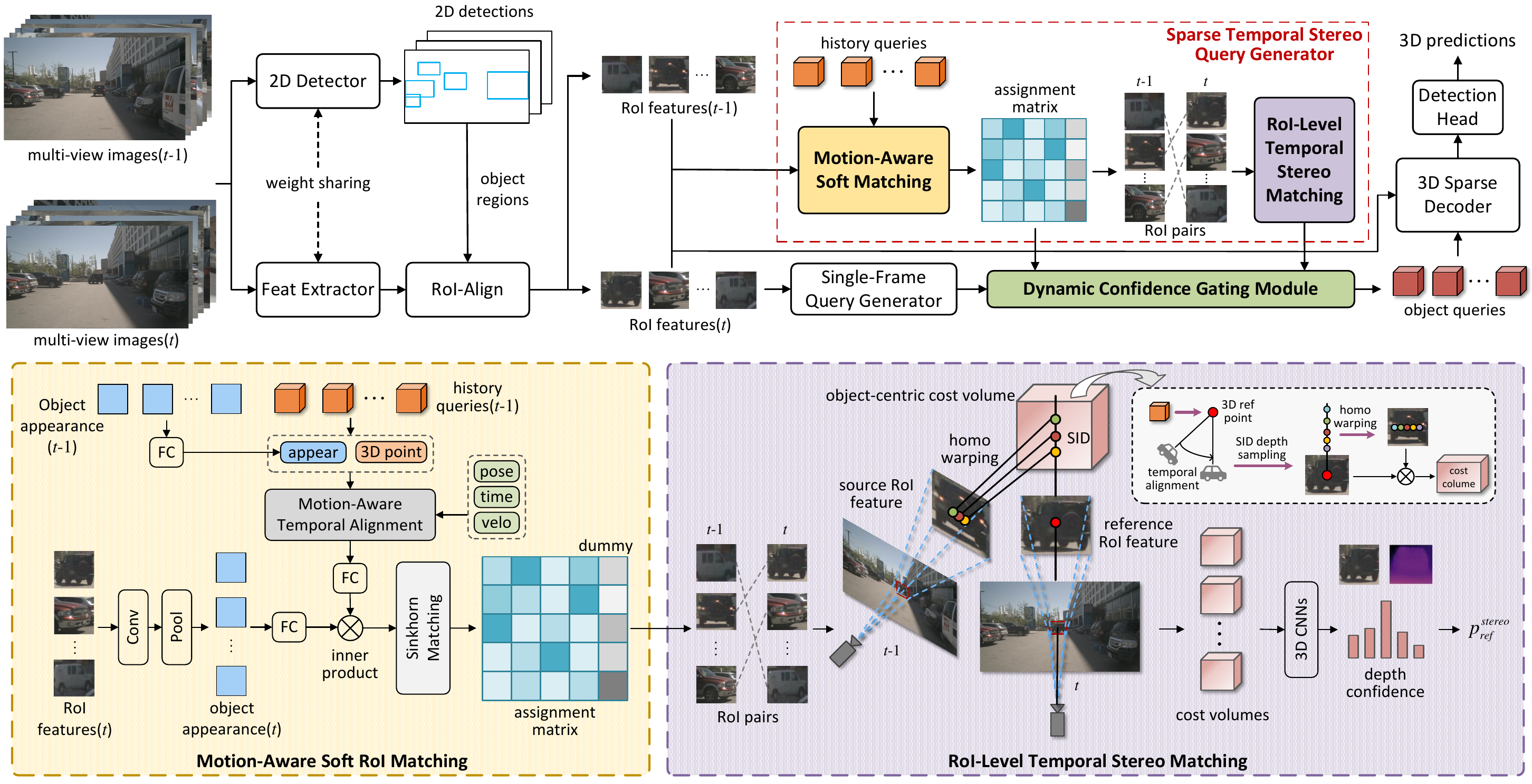}
\caption{Overall architecture of StereoMV2D. Given multi-view images from adjacent timestamps, the model first extracts image features and obtains RoI features corresponding to 2D detections. These RoI features, together with the historical queries, are fed into the sparse temporal stereo query generator. The generator begins by performing motion-aware soft matching to compute a matching matrix between objects across frames. After identifying RoI pairs from adjacent timestamps, an RoI-level temporal stereo matching module constructs cost volumes between all matched RoI pairs to predict object depths. To handle newly appearing objects and occlusion cases, we retain the monocular implicit query generator and integrate the 3D reference point proposals from the monocular and stereo branches using a dynamic confidence gating strategy. The fused depth-aware queries, equipped with strong positional priors, are then refined through interactions with RoI features via a sparse decoder, and finally passed through the detection head to produce 3D predictions.}
\label{fig_2}
\end{figure*}

The overall pipeline of StereoMV2D follows the design of MV2D, as illustrated in Fig. 2, and consists of an image feature extractor, a 2D detector, a query generator, a sparse decoder, and a detection head, while taking multi-view images from adjacent timestamps as input. Given $N$ multi-view images at the current time $t$ and the previous time $t-1$, the model first extracts image features using an image backbone and obtains 2D detections using a 2D detector (e.g., Faster R-CNN\cite{ren2016faster}). From these detections, we acquire 2D bounding boxes and extract the corresponding RoI features via RoI-Align. With RoI features from the two timestamps, we generate candidate 3D reference points using both a single-frame query generator and a sparse temporal-stereo query generator. The single-frame generator directly follows MV2D’s dynamic object query generator, which leverages per-RoI image features together with the RoI-specific equivalent intrinsic matrix to implicitly encode the object’s potential 3D location from a monocular perspective. In parallel, the temporal-stereo generator integrates explicit geometric cues across adjacent frames and is composed of two modules: a Motion-Aware Soft Matching (MASM) module that establishes differentiable inter-frame RoI associations, and an RoI-level Temporal Stereo Matching (RTSM) module that constructs lightweight object-centric cost volumes to infer stereo-based depth priors.

The outputs of the two query generators, together with the matching matrix and RoI features, are fed into a dynamic confidence gating module, which learns to assess the reliability of the monocular and stereo branches under different conditions and produces a weighted fusion of their 3D reference proposals as the positional prior for query initialization. The resulting object queries aggregate features from RoIs through the sparse decoder, and the final 3D predictions are generated by the detection head based on the updated queries. Detailed descriptions are provided in the following sections.

\subsection{Sparse Temporal Stereo Query Generator}
Leveraging reliable and mature 2D detectors provides high-quality object locations and bounding boxes, which serve as a foundation for exploring the geometric consistency of objects across adjacent frames within confined image regions. Our temporal stereo query generator adopts a two-stage design: it first identifies object correspondences through Motion-Aware Soft Matching, and then performs RoI-Level Temporal Stereo Matching within paired RoIs to construct lightweight, object-focused cost volumes. This design enables StereoMV2D to selectively incorporate stereo cues only where meaningful object evidence exists, yielding accurate and robust depth priors for downstream sparse decoding.

\paragraph{Motion-Aware Soft Matching (MASM)}
To perform temporal stereo on the same object across adjacent timestamps, it is essential to first establish one-to-one correspondences between 2D detections from the two frames. This process is analogous to data association between detections and tracks in the multi-object tracking (MOT) literature\cite{bergmann2019tracking, aharon2022bot}. Inspired by this, we propose a Motion-Aware Soft Matching module highlighted in the yellow block of Fig. 2. Instead of relying solely on appearance similarity in the image domain, incorporating motion cues in 3D space provides significantly more stability for 3D perception tasks. Specifically, given the multi-view image features from two adjacent timestamps $\textbf{F}_j=\{F_j^{i}\}_{i=1}^{N}, j\in\{t-1,t\}$ and their corresponding 2D detection bounding boxes $\textbf{B}_j=\{B_{j}^{i}\}_{i=1}^{N}$, we first extract the RoI features $\textbf{R}_j$ of each object using RoI-Align:
\begin{equation}
\begin{array}{l}
\textbf{R}_j=\text{RoI-Align}(\textbf{F}_j,\textbf{B}_j)
\end{array}
\tag{1}.
\end{equation}
Here, $\textbf{R}_j\in\mathbb{R}^{M_j\times H^{roi}\times W^{roi}\times C}$, where $M_j$ denotes the total number of 2D detections across all camera views at time $j$, $H^{roi}$ and $W^{roi}$ represent the fixed RoI height and width, and $C$ denotes the number of feature channels. We then encode each RoI feature into an object appearance embedding $\textbf{O}_j\in\mathbb{R}^{M_j\times C}$ using a lightweight network composed of convolution and average pooling layers.

To model the temporal motion of each object, we store the historical queries from the previous timestamp in a memory buffer and decode from them the associated 3D reference points $Q^p_{t-1}\in \mathbb{R}^{M_{t-1}\times3}$ and velocities $Q^v_{t-1}\in \mathbb{R}^{M_{t-1}\times2}$ (within BEV plane). Following \cite{wang2023exploring}, we employ motion-aware layer normalization to model object motion patterns and align historical information to the current frame. Since the ego movement over time, we first align the reference positions from the previous timestamp to the current frame by ego transformation. Given the ego-pose matrices $\boldsymbol{E}_{t-1}$ and $\boldsymbol{E}_t$ at the two timestamps, the ego-motion transformation from time $t-1$ to $t$ is computed as:
\begin{equation}
\begin{array}{l}
\boldsymbol{E}_{t-1}^t=\boldsymbol{E}_t^{inv}\cdot \boldsymbol{E}_{t-1}
\end{array}
\tag{2}.
\end{equation}
We then assume that all objects remain static in the world coordinate system and align all historical 3D reference points $Q^p_{t-1}$ to the current timestamp as:
\begin{equation}
\begin{array}{l}
\hat{Q}^p_{t}=\boldsymbol{E}_{t-1}^t\cdot Q^p_{t-1}
\end{array}
\tag{3},
\end{equation}
where $\hat{Q}^p_{t}$ is the aligned 3D points. With the historical query velocities $Q_{t-1}^v$, the inter-frame time interval $\Delta t$, and the ego-motion transformation $\boldsymbol{E}_{t-1}^t$, we compute the affine parameters $\gamma$ and $\beta$ through two linear layers as follows:
\begin{equation}
\begin{array}{l}
\gamma=\text{Linear}_1(\text{PE}[Q_{t-1}^v, \Delta t, \boldsymbol{E}_{t-1}^t])\\
\beta=\text{Linear}_2(\text{PE}[Q_{t-1}^v, \Delta t, \boldsymbol{E}_{t-1}^t])
\end{array}
\tag{4},
\end{equation}
where $[\cdot]$ is the concatenate operator and $\text{PE}(\cdot)$ is a positional encoding function. These parameters are then applied to perform affine transformations on the aligned reference points $\hat{Q}^p_{t}$ and the historical object appearance $O_{t-1}$ to:
\begin{equation}
\begin{array}{cc}
\hat{Q}^{pe}_{t}=\gamma\cdot\text{LN}(\text{MLP}(\hat{Q}^p_{t}))+\beta\\
\hat{O}_{t}=\gamma\cdot\text{LN}(O_{t-1})+\beta
\end{array}
\tag{5},
\end{equation}
where $\hat{Q}^{pe}_{t}, \hat{O}_{t}\in \mathbb{R}^{M_{t-1}\times C}$ denote the motion-aware positional encoding and motion-aware appearance embedding, respectively. Here, $\text{MLP}(\cdot)$ represents a multilayer perceptron that maps 3D points to positional encodings, and $\text{LN}(\cdot)$ denotes layer normalization. Then we sum the two components along the channel dimension to produce the temporal-aligned historical object embedding $\hat{Q}_t$:
\begin{equation}
\begin{array}{l}
\hat{Q}_t=\hat{Q}^{pe}_{t}+\hat{O}_{t}
\end{array}
\tag{6}.
\end{equation}

After obtaining the temporal-aligned historical object embedding $\hat{Q}_t$ and the current object embedding $O_t$, we compute their matching relationship. Before performing the matching, we append an additional all-zero feature vector to $\hat{Q}_t$. This vector functions as a dummy column in the subsequent matching matrix, allowing newly appeared objects at time $t$ to be matched without disrupting the one-to-one associations among existing objects. To ensure the two embeddings lie in a comparable feature space, we apply two linear projection layers to map $\hat{Q}_t$ and $O_t$ into a shared latent space with the dimension of $C'$, producing $\hat{Q}'_t\in \mathbb{R}^{M_{t-1}\times C'}$ and $O'_t\in \mathbb{R}^{M_{t}\times C'}$. We then compute an initial similarity matrix using the scaled dot-product: 
\begin{equation}
\begin{array}{l}
\boldsymbol{S}_t=\frac{O'_t(\hat{Q}'_t)^\mathrm{T}}{\sqrt{C'}} \in \mathbb{R}^{M_t\times (M_{t-1}+1)}
\end{array}
\tag{7}.
\end{equation}
This similarity matrix is subsequently refined using Sinkhorn-Knopp algorithm for differentiability, which performs entropic regularization and iterative row–column normalization to obtain a doubly stochastic assignment matrix:
\begin{equation}
\begin{array}{l}
\boldsymbol{A}_t=\text{Sinkhorn}(\boldsymbol{S}_t)
\end{array}
\tag{8}.
\end{equation}
In this matrix, each row corresponds to an object at the current timestamp, and each column corresponds to a historical object or the dummy column, where the last column explicitly represents “newly appeared objects.” This soft matching mechanism preserves an approximate one-to-one assignment while allowing flexibility for unmatched or newly emerged objects, thereby providing a stable and differentiable object association prior for subsequent temporal stereo matching.

\paragraph{RoI-Level Temporal Stereo Matching (RTSM)}
With the motion-aware soft matching module producing a differentiable association matrix $A_t$, we obtain reliable RoI correspondences between the current and previous frames. For each matched pair, the RoI feature at time $t$ is treated as the reference feature, while the corresponding RoI feature at time $t-1$ serves as the source feature for stereo reasoning as shown in the purple block of Fig. 2. Before constructing the cost volume, we must first determine an appropriate depth sampling range. 

To efficiently constrain the sampling space, we leverage the temporal alignment performed in the previous stage: each row of $\boldsymbol{A}_t$ is used as a set of weights to compute a weighted sum over the aligned historical 3D reference points $\hat{Q}^p_{t}$, yielding a prior depth center for each current-frame RoI. Specifically, for each current-frame RoI $R_t^m, m=1,\ldots,M_t$, we compute a prior 3D center $p_m$ by performing a weighted aggregation as follow:
\begin{equation}
\begin{array}{l}
p_m=\sum\limits_{n=1}^{M_{t-1}}\boldsymbol{A}_t(m,n)\hat{Q}^p_{t,n}
\end{array}
\tag{9},
\end{equation}
where $\hat{Q}^p_{t,n}$ is the $n$-th aligned history 3D point in $\hat{Q}^p_{t}$. For the subsequent coordinate transformation, we convert it into the homogeneous form: $\bar{p}_m=[p_m|1]^\mathrm{T}$. Then we transform it into the coordinate system of the $m$-th RoI $R_t^m$. Due to the rescaling operation in RoIAlign, the geometric projection relationship of the original camera is no longer preserved. Following \cite{wang2023object}, the rescaling performed on different RoIs can be interpreted as a perspective projection under equivalent intrinsic matrices specific to each RoI. Formally, let $K_i$ denotes the original camera intrinsic matrix of the $i$-th camera view where $R_t^m$ is located:
\begin{equation}
\begin{array}{l}
\boldsymbol{K}_i=\begin{bmatrix}
f_x & 0 & o_x & 0\\
0 & f_y & o_y & 0\\
0 & 0 & 1 & 0\\
0 & 0 & 0 & 1
\end{bmatrix}
\end{array}
\tag{10}.
\end{equation}
Given the 2D detection bounding box $B_{t,m}^i=(x_{min}^m, y_{min}^m, x_{max}^m, y_{max}^m)$ that generates $R_t^m$, then the equivalent camera intrinsic matrix of $R_t^m$ is formulated as:
\begin{equation}
\begin{array}{l}
\boldsymbol{K}_i^m=\begin{bmatrix}
f_x*r_x & 0 & (o_x-x_{min}^m)*r_x & 0\\
0 & f_y*r_y & (o_y-y_{min}^m)*r_y & 0\\
0 & 0 & 1 & 0\\
0 & 0 & 0 & 1
\end{bmatrix}
\end{array}
\tag{11},
\end{equation}
where $r_x=W^{roi}/(x_{max}^m-x_{min}^m)$, $r_y=H^{roi}/(y_{max}^m-y_{min}^m)$. Let $\boldsymbol{T_i}\in \mathbb{R}^{4\times 4}$ represent the camera extrinsic matrix from world coordinates to the $i$-th camera view. The prior 3D center can then be projected into the coordinate system of the $m$-th RoI as:
\begin{equation}
\begin{array}{l}
\tilde{p}_m=\boldsymbol{K}_i^m\cdot\boldsymbol{T}_i\cdot\bar{p}_m
\end{array}
\tag{12},
\end{equation}
where $\tilde{p}_m=(u_m*d_m,v_m*d_m,d_m,1)^\mathrm{T}$ is a homogeneous 2.5D coordinate of the $m$-th RoI. Here, $d_m$ is the depth of the prior 3D center, and $(u_m,v_m)$ denote its 2D coordinates in the rescaled RoI coordinate frame. 

Based on this prior depth $d_m$, we adaptively constructs a depth sampling interval to constrain the computational range for temporal stereo. Specifically, given a scaling factor $\alpha>0$, we define the depth interval boundaries as $[d_m^{min},d_m^{max}], d_m^{min}=d_m/\alpha, d_m^{max}=\alpha\cdot d_m$. Within this range, we adopt Spacing-Increasing Discretization (SID)\cite{wang2022sts} to sample $D$ depth candidates in log space as:
\begin{equation}
\begin{array}{l}
d_{m,k}=d_m^{min}(d_m^{max}\big /d_m^{min})^{\frac{k}{D-1}}, k=0,\cdots,D-1
\end{array}
\tag{13}.
\end{equation}
This spacing-increasing scheme allocates denser samples at smaller depths and sparser samples at larger depths, which helps avoid geometric misalignment for nearby objects while keeping the overall sampling cost low. With the depth candidates $\{d_{m,k}\}_{k=0}^{D-1}$, each depth value is combined with every pixel location $(u_l,v_l)$ in the RoI patch (with $H^{roi}\times W^{roi}$ pixels) to form a set of 2.5D points $p_{m,l,k}=(u_l,v_l,d_{m,k})$.

To obtain their correspondences in the source RoI at time $t-1$, each candidate point is projected through a homography induced by the camera geometry and temporal motion. Importantly, the reference-frame RoI and its matched source RoI may originate from different camera views, because an object may not remain in the same physical camera between frames. Nevertheless, MASM step produces reliable temporal associations, enabling RTSM to reconstruct consistent cross-frame correspondences. Denote by $\boldsymbol{K}_i^m$ and $\boldsymbol{K}_{i'}^n$ the equivalent intrinsic matrices of the reference and source RoIs, and by $\boldsymbol{M}_{ref2src}$ the $4\times 4$ transformation from the reference camera at time $t$ to the source camera at time $t-1$. The warping operation for each 2.5D candidate point can be written compactly as:
\begin{equation}
\begin{array}{l}
p_{m,l,k}^{src}=\boldsymbol{K}_{i'}^n \boldsymbol{M}_{ref2src}(\boldsymbol{K}_i^m)^{-1}\begin{bmatrix}
u_l d_{m,k}\\
v_l d_{m,k}\\
d_{m,k}\\
1
\end{bmatrix}
\end{array}
\tag{14},
\end{equation}
where $p_{m,l,k}^{src}$ is the homogeneous projection on the source RoI. Using these warped coordinates, the reference-frame RoI features collect the corresponding source-frame features through differentiable sampling. The similarity between them is computed via an inner product to construct a lightweight, object-centric cost volume. A subsequent 3D CNN processes this cost volume to estimate a depth-confidence distribution for the RoI $R_t^m$. We then select, for each pixel, the depth candidate with the highest confidence and average these depths over the entire RoI. Together with the RoI’s center pixel location, this yields a 2.5D reference center $p_m^{center}$. Finally, this 2.5D reference point is lifted back into 3D world coordinates using $\boldsymbol{K}_i^m$ and $\boldsymbol{T}_i$. The resulting stereo-based 3D reference point for the query is:
\begin{equation}
\begin{array}{l}
p_{ref,m}^{stereo}=(\boldsymbol{T}_i)^{-1}\cdot(\boldsymbol{K}_i^m)^{-1}\cdot p^{center}_m
\end{array}
\tag{15}.
\end{equation}
This 3D position serves as the geometrically consistent reference point for query initialization.

\subsection{Dynamic Confidence Gating Module}
For each current-frame RoI $R_m^t$, the sparse temporal stereo query generator and the single-frame implicit query generator respectively produce two candidate 3D reference points, denoted as $p_{ref,m}^{stereo}$ and $p_{ref,m}^{mono}$. However, due to object occlusion or new object emergence, a current RoI may not always find a valid correspondence in the previous frame. Although the assignment matrix incorporates a dummy column to accommodate such cases, the historical position associated with the dummy column is fictitious and inevitably introduces uncertainty. In these situations, we expect the model to compensate for this ambiguity via the monocular branch. Therefore, we learn a confidence score $c_m\in[0,1]$ that indicates the degree to which the temporal stereo prior should be trusted. Based on this confidence, the final positional prior used for query initialization is dynamically fused as:
\begin{equation}
\begin{array}{l}
p_{ref,m}=c_m p_{ref,m}^{stereo}+(1-c_m)p_{ref,m}^{mono}
\end{array}
\tag{16}.
\end{equation}

Let $\boldsymbol{A}_t\in \mathbb{R}^{M_t\times (M_{t-1}+1)}$ denote the soft association matrix between current-frame RoIs and previous-frame RoIs augmented with a dummy column, and $\boldsymbol{A}_t [m,n]$ is the association probability of current $R_t^m$ and to previous RoI $R_{t-1}^n$. From the $m$-th row we extract the real mass
\begin{equation}
\begin{array}{l}
\rho_m^{real}=\sum\limits_{n=1}^{M_{t-1}}\boldsymbol{A}_t[m,n]
\end{array}
\tag{17},
\end{equation}
which measures how much the current RoI is explained by non-dummy tracks, and the dummy mass $\rho_m^{dum}=\boldsymbol{A}_t[m,M_{t-1}+1]$. From the real columns $\{1,\ldots,M_{t+1}\}$ of $\boldsymbol{A}_t[m,\cdot]$ we extract three distributional statistics that correlate with stereo reliability: the peak association over real columns $\rho_m^{max}=\text{max}_{1\leq n\leq M_{t-1}}\boldsymbol{A}_t[m,n]$, the gap between top-1 and top-2 assignments $\rho_m^{gap}=\boldsymbol{A}_t^{(1)}[m]-\boldsymbol{A}_t^{(2)}[m]$ and the entropy
\begin{equation}
\begin{array}{l}
H_m=-\sum\limits_{n=1}^{M_{t-1}}\boldsymbol{A}_t[m,n]\text{log}(\boldsymbol{A}_t[m,n]+\epsilon)
\end{array}
\tag{18},
\end{equation}
where $\epsilon$ is a constant of numerical stability. Intuitively, large $\rho_m^{max}$ and $\rho_m^{gap}$, and small $\rho_m^{dum}$ and $H_m$ indicate a sharp, unambiguous association with a real previous RoI, hence a more reliable stereo prior. To complement the association shape, we assess the agreement between the current RoI feature and its most likely past counterpart. We define a lightweight embedding $\psi(\cdot)$ by global average pooling followed by a linear projection and layer normalization. With $n^*=\text{argmax}_{1\leq n\leq M_{t-1}}\boldsymbol{A}_t [m,n]$, the cosine affinity is
\begin{equation}
\begin{array}{l}
\phi_m=\frac{\langle\psi(R_t^m),\psi(R_{t-1}^{n^*})\rangle}{\Vert\langle\psi(R_t^m)\Vert\Vert\psi(R_{t-1}^{n^*})\Vert}
\end{array}
\tag{19},
\end{equation}
which serves as a supplementary signal for appearance consistency: When the appearance is consistent and the correlation is sharp, there is a tendency to trust the temporal branch.

We assemble a five-dimensional descriptor $z_m=[1-\rho_m^{dum},\rho_m^{max},\rho_m^{gap},-H_m,\phi_m]\in\mathbb{R}^5$ and predict the gate via a two-layer MLP $f_{\theta}:\mathbb{R}^5\rightarrow\mathbb{R}$ and a sigmoid function $\sigma(\cdot)$:
\begin{equation}
\begin{array}{l}
c_m=
\begin{cases}
\sigma(f_{\theta}(z_m)), & \rho_m^{real}>\epsilon\\
0, & \text{otherwise}
\end{cases}
\end{array}
\tag{20}.
\end{equation}
When the association carries negligible real mass, i.e., $\rho_m^{real}\leq\epsilon$, we force a safe fallback to the monocular prior. This rule handles newly appeared or heavily occluded objects without requiring any heuristic thresholds on the stereo branch itself. 

\subsection{3D Sparse Decoder}
After obtaining the object queries initialized with the fused positional priors, we employ a 3D sparse decoder to perform iterative feature refinement. The decoder follows the standard Transformer architecture used in DETR\cite{carion2020end}, consisting of $l$ stacked decoder layers. Each layer is composed of a self-attention module, a cross-attention module, and feed-forward components. Unlike PETR\cite{liu2022petr} and other approaches that perform dense cross-attention over full multi-view image features, our framework adopts the sparse cross-attention strategy introduced in MV2D: each query interacts only with the RoI features corresponding to its associated 2D detection. This targeted interaction eliminates redundant computation on background regions—functionally similar to token sparsification—thereby enabling efficient and object-centric feature aggregation. Following MV2D, the decoder leverages all input frames’ camera extrinsic matrices as well as the ego-motion transformation to enable cross-frame RoI interaction, allowing the updated queries to incorporate multi-view and temporal cues simultaneously. Finally, the refined queries are passed through task-specific MLP heads for classification and 3D bounding box regression, producing the final 3D detection results.

\subsection{Overall Loss}
The 2D detector and the 3D detector in our framework are jointly trained and share the same image backbone. The overall training objective consists of both 2D detection loss $L_{2d}$ and 3D detection loss $L_{3d}$. For the 2D detection task, the loss is composed of a classification loss $L_{cls2d}$ and a bounding-box regression loss $L_{reg2d}$, implemented using cross-entropy loss and L1 loss, respectively. For the 3D detection task, the classification loss $L_{cls3d}$ is formulated using the focal loss to address class imbalance, while the regression loss $L_{reg3d}$ adopts the L1 loss to supervise the predicted 3D bounding box parameters. The overall loss is computed as: 
\begin{equation}
\begin{array}{l}
L_{total}=L_{cls2d}+L_{reg2d}+\lambda_{1}L_{cls3d}+\lambda_{2}L_{reg3d}
\end{array}
\tag{21},
\end{equation}
where $\lambda_{1}$ and $\lambda_{2}$ are weighting coefficients that balance the contributions of the 3D classification and regression terms. 

\section{Experiments}
\subsection{Datasets and Metrics}
\textbf{The nuScenes Dataset} The nuScenes dataset \cite{caesar2020nuscenes} provides a full 360° surround-view using six cameras (1600×900, 12 Hz) and a 32-beam LiDAR (20 Hz). It contains 1,000 driving scenes of 20 seconds each, collected under diverse weather, lighting, and geographic environments. Keyframes are annotated at 2 Hz with 3D bounding boxes for 10 object categories. Following the official split, 28,130 frames from 750 scenes are used for training, while 6,019 and 6,008 frames are reserved for validation and testing. We follow the official benchmark to evaluate 3D detection performance using mean Average Precision (mAP), the nuScenes Detection Score (NDS), and several True Positive (TP) quality metrics, including mean Average Translation Error (mATE), mean Average Scale Error (mASE), mean Average Orientation Error (mAOE), mean Average Velocity Error (mAVE) and mean Average Attribute Error (mAAE). Among these, mAP and NDS serve as the primary evaluation metrics. 

\textbf{The Argoverse 2 Dataset} The Argoverse 2 (AR2) Sensor dataset \cite{wilson2023argoverse} provides a large-scale, diverse collection of urban driving scenes captured using multiple synchronized cameras (2048×1550, 20Hz). It contains 1,000 driving sequences, of which 700 are used for training, 150 for validation, and 150 for testing. The dataset covers a wide range of environments, including urban centers, suburban roads, and highway scenes, and annotates 26 object categories with high-quality 3D bounding boxes. Each sequence is recorded at high frame rates with full-surround camera coverage, offering rich visual information for multi-view 3D perception. Following the official evaluation protocol, we assess performance using the AR2 3D detection metrics, including mAP, mATE, mASE, and mAOE, along with the Composite Detection Score (CDS) that aggregates multiple error terms into a single benchmark metric.

\subsection{Implementation Details}
Our experimental settings follow those of the baseline for fair comparison. StereoMV2D adopts Faster R-CNN\cite{ren2016faster} as the 2D detector, initialized with nuImages-pretrained \cite{caesar2020nuscenes} weights. We employ ResNet-50, ResNet-101, and VoVNetV2\cite{lee2019energy} as the image backbones. For ResNet-50 and ResNet-101, deformable convolutions\cite{zhu2020deformable} are used in stages 3 and 4, and an FPN is applied to generate five multi-scale feature levels. Based on the number of frames involved in sparse decoder feature interaction, we construct two configurations: StereoMV2D-S, which interacts only with RoI features from the current frame, and StereoMV2D-T, which incorporates RoI features from both the current and previous frames. VoVNetV2 is initialized with DD3D-pretrained weights\cite{park2021pseudo}. The decoder contains $l=6$ layers. For the nuScenes dataset, StereoMV2D-S with a ResNet-50 backbone uses an input resolution of 1408 × 512, while StereoMV2D-T uses 1600 × 640. All configurations based on ResNet-101 and VoVNetV2 adopt an input resolution of 1600 × 640. The model is trained for 72 epochs on the nuScenes training set without CBGS. For the AR2 dataset, the input resolution is 960 × 640. We use the nuImages pre-trained weights for 2D detector and the whole model is trained for 6 epochs. We use the AdamW optimizer with cosine annealing scheduling, an initial learning rate of $2\times10^{-4}$, and a weight decay of 0.01. All experiments are conducted on 4 NVIDIA L20 GPUs.

For the other hyperparameter settings used in the algorithm, both the height $H^{roi}$ and $W^{roi}$ of each RoI are set to 7. We use a small constant $\epsilon=1\times10^{-6}$ for numerical stability. The weighting coefficients of the overall loss function are set to $\lambda_{1}=0.2$ and $\lambda_{2}=0.025$.

\subsection{Comparison with Existing Methods}
\begin{table*}[t]
  \centering
  \caption{3D object detection results on nuScenes $val$ set. \dag: Pre-trained from FCOS3D. *: Pre-trained on nuImages. FPS is measured on RTX3090 with fp32. Our model is marked in gray . Bold indicates the best performance.}
  \label{tab:1}
  \setlength{\tabcolsep}{2mm}{
  \begin{tabular}{c|cc|c|cc|ccccc|c}
  \toprule
  Method & Img Size & Backbone & Frames & mAP$\uparrow$ & NDS$\uparrow$ & mATE$\downarrow$ & mASE$\downarrow$ & mAOE$\downarrow$ & mAVE$\downarrow$ & mAAE$\downarrow$ & FPS$\uparrow$ \\
  \midrule
  DETR3D\cite{wang2022detr3d} & $1600\times900$ & ResNet-50 & 1 & 0.302 & 0.373 & 0.811 & 0.282 & 0.493 & 0.979 & 0.212 & 6.2 \\
  PETR\cite{liu2022petr} & $1408\times512$ & ResNet-50 & 1 & 0.339 & 0.403 & 0.748 & 0.273 & 0.539 & 0.907 & 0.203 & 9.7 \\
  PETRv2\cite{liu2023petrv2} & $1600\times640$ & ResNet-50 & 2 & 0.398 & 0.494 & 0.690 & 0.273 & 0.467 & 0.424 & 0.195 & - \\
  MV2D-S*\cite{wang2023object} & $1408\times512$ & ResNet-50 & 1 & 0.398 & 0.440 & 0.665 & 0.269 & 0.507 & 0.946 & 0.203 & 6.2/25.3 \\
  MV2D-T*\cite{wang2023object} & $1600\times640$ & ResNet-50 & 2 & 0.459 & 0.546 & 0.613 & 0.265 & 0.388 & 0.385 & 0.179 & 3.1/14.9 \\
  \rowcolor{gray!30}StereoMV2D-S* & $1408\times512$ & ResNet-50 & 2 & 0.417 & 0.461 & 0.640 & 0.268 & 0.463 & 0.742 & 0.187 & 3.9/19.4 \\
  \rowcolor{gray!30}StereoMV2D-T* & $1600\times640$ & ResNet-50 & 2 & \textbf{0.481} & \textbf{0.569} & \textbf{0.591} & \textbf{0.261} & \textbf{0.352} & \textbf{0.308} & \textbf{0.153} & 2.2/11.3 \\
  \midrule
  DETR3D\dag\cite{wang2022detr3d} & $1600\times900$ & ResNet-101 & 1 & 0.349 & 0.434 & 0.716 & 0.268 & 0.379 & 0.842 & 0.200 & 3.7 \\
  BEVFormer\dag\cite{li2024bevformer} & $1600\times900$ & ResNet-101 & 4 & 0.416 & 0.517 & 0.673 & 0.274 & 0.372 & 0.394 & 0.198 & 3.0 \\
  OA-BEVFormer\dag\cite{chu2025oa} & $1600\times900$ & ResNet-101 & 4 & 0.431 & 0.528 & 0.664 & 0.272 & 0.388 & 0.345 & 0.205 & - \\
  HV-BEV\dag\cite{wu2025hv} & $1600\times900$ & ResNet-101 & 4 & 0.439 & 0.533 & 0.617 & 0.264 & 0.388 & 0.375 & \textbf{0.127} & 2.2 \\
  PETRv2\dag\cite{liu2023petrv2} & $1600\times640$ & ResNet-101 & 2 & 0.421 & 0.524 & 0.681 & 0.267 & 0.357 & 0.377 & 0.186 & - \\
  BEVDet4D\cite{huang2022bevdet4d} & $1600\times640$ & Swin-T & 2 & 0.421 & 0.545 & 0.579 & \textbf{0.258} & 0.329 & 0.301 & 0.191 & 1.9 \\
  GraphDETR4D\dag\cite{chen2024graph} & $1408\times512$ & ResNet-101 & 2 & 0.462& 0.577 & - & - & - & - & - & - \\
  Sparse4D\dag\cite{lin2022sparse4d} & $1600\times900$ & ResNet-101 & 9 & 0.444 & 0.550 & 0.603 & 0.276 & 0.360 & 0.309 & 0.178 & 4.3 \\
  WidthFormer\dag\cite{yang2024widthformer} & $1408\times512$ & ResNet-101 & 2 & 0.423 & 0.531 & 0.609 & 0.269 & 0.412 & 0.302 & 0.210 & - \\
  Fast-BEV\dag\cite{li2024fast} & $1600\times900$ & ResNet-101 & 4 & 0.413 & 0.535 & 0.584 & 0.279 & \textbf{0.311} & 0.329 & 0.206 & 11.6 \\
  SOLOFusion\cite{park2022time} & $1408\times512$ & ResNet-101 & 16+1 & 0.483 & 0.582 & \textbf{0.503} & 0.264 & 0.381 & 0.246 & 0.207 & - \\
  MV2D-S*\cite{wang2023object} & $1600\times640$ & ResNet-101 & 1 & 0.424 & 0.470 & 0.654 & 0.267 & 0.416 & 0.888 & 0.200 & 4.5/17.7 \\
  MV2D-T*\cite{wang2023object} & $1600\times640$ & ResNet-101 & 2 & 0.471 & 0.561 & 0.593 & 0.262 & 0.340 & 0.368 & 0.184 & 2.3/10.7 \\
  \rowcolor{gray!30}StereoMV2D-S* & $1600\times640$ & ResNet-101 & 2 & 0.442 & 0.497 & 0.621 & 0.264 & 0.368 & 0.524 & 0.195 & 3.0/13.8 \\
  \rowcolor{gray!30}StereoMV2D-T* & $1600\times640$ & ResNet-101 & 2 & \textbf{0.499} & \textbf{0.592} & 0.535 & \textbf{0.258} & 0.317 & \textbf{0.239} & 0.171 & 1.6/8.4 \\
  \bottomrule
  \end{tabular} }
\end{table*}

\begin{table*}[t]
  \centering
  \caption{3D detection results on nuScenes $test$ set. \dag: Pre-trained from FCOS3D. \ddag: Pre-trained from DD3D. *: Pre-trained on nuImages. Our model is marked in gray . Bold indicates the best performance.}
  \label{tab:2}
  \setlength{\tabcolsep}{2.6mm}{
  \begin{tabular}{c|cc|c|cc|ccccc}
  \toprule
  Method & Img Size & Backbone & Frames & mAP$\uparrow$ & NDS$\uparrow$ & mATE$\downarrow$ & mASE$\downarrow$ & mAOE$\downarrow$ & mAVE$\downarrow$ & mAAE$\downarrow$  \\
  \midrule
  BEVDet4D\cite{huang2022bevdet4d} & $1600\times640$ & Swin-T & 2 & 0.451 & 0.569 & \textbf{0.511} & \textbf{0.241} & 0.386 & \textbf{0.301} & 0.121 \\
  GraphDETR3D\dag\cite{chen2022graph} & $1600\times640$ & ResNet101 & 1 & 0.418 & 0.472 & 0.668 & 0.250 & 0.440 & 0.876 & 0.139 \\
  PETRv2\dag\cite{liu2023petrv2} & $1600\times640$ & ResNet101 & 2 & 0.456 & 0.553 & 0.601 & 0.249 & 0.391 & 0.382 & 0.123 \\
  HV-BEV\dag & $1600\times900$ & ResNet101 & 4 & 0.464 & 0.556 & 0.604 & 0.261 & 0.380 & 0.393 & 0.132 \\
  MV2D-T*\cite{wang2023object} & $1600\times640$ & ResNet101 & 2 & 0.483 & 0.573 & 0.567 & 0.249 & 0.359 & 0.395 & \textbf{0.116}\\
  \rowcolor{gray!30}StereoMV2D-T* & $1600\times640$ & ResNet101 & 2 & \textbf{0.501} & \textbf{0.588} & 0.551 & 0.244 & \textbf{0.357} & 0.366 & \textbf{0.116}\\
  \midrule
  DETR3D\ddag\cite{wang2022detr3d} & $1600\times900$ & VoVNetV2 & 1 & 0.412 & 0.479 & 0.641 & 0.255 & 0.394 & 0.845 & 0.133 \\
  GraphDETR3D\ddag\cite{chen2022graph} & $1600\times640$ & VoVNetV2 & 1 & 0.425 & 0.495 & 0.621 & 0.251 & 0.386 & 0.790 & 0.128 \\
  PETRv2\ddag\cite{liu2023petrv2} & $1600\times640$ & VoVNetV2 & 2 & 0.490 & 0.582 & 0.561 & 0.243 & 0.361 & 0.343 & 0.120 \\
  Sparse4D\dag\cite{lin2022sparse4d} & $1600\times900$ & VoVNetV2 & 9 & 0.511 & 0.595 & 0.533 & 0.263 & 0.369 & \textbf{0.317} & 0.124 \\
  VEDet\ddag\cite{chen2023viewpoint} & $1600\times640$ & VoVNetV2 & 1 & 0.505 & 0.585 & 0.545 & 0.244 & 0.346 & 0.421 & 0.123 \\
  HV-BEV\ddag\cite{wu2025hv} & $1600\times900$ & VoVNetV2 & 4 & 0.505 & 0.598 & 0.544 & 0.249 & 0.353 & 0.318 & 0.117 \\
  OA-BEVFormer\ddag\cite{chu2025oa} & $1600\times900$ & VoVNetV2 & 4 & 0.494 & 0.575 & 0.574 & 0.256 & 0.377 & 0.385 & 0.132 \\
  BEVStereo\ddag\cite{li2023bevstereo} & $1600\times640$ & VoVNetV2 & 2 & 0.525 & 0.610 & \textbf{0.431} & 0.246 & 0.358 & 0.357 & 0.138 \\
  MV2D-T\ddag\cite{wang2023object} & $1600\times640$ & VoVNetV2 & 2 & 0.511 & 0.596 & 0.525 & 0.243 & 0.357 & 0.357 & 0.120 \\
  \rowcolor{gray!30}StereoMV2D-T\ddag & $1600\times640$ & VoVNetV2 & 2 & \textbf{0.535} & \textbf{0.622} & 0.437 & \textbf{0.241} & \textbf{0.338} & 0.323 & \textbf{0.113} \\ 
  \bottomrule
  \end{tabular} }
\end{table*}

We conduct a comprehensive comparison between StereoMV2D and a range of existing state-of-the-art methods. Tab. \uppercase\expandafter{\romannumeral1}and Tab. \uppercase\expandafter{\romannumeral2} present performance comparisons on the nuScenes validation set and test set, respectively. Tab. \uppercase\expandafter{\romannumeral3} reports the performance comparison on the Argoverse 2 validation set. 

As shown in Tab. \uppercase\expandafter{\romannumeral1}, our StereoMV2D consistently surpasses the baseline MV2D across all backbone choices and both S/T configurations, demonstrating the effectiveness of incorporating sparse temporal stereo into the query generation process. The best model achieves 49.9\% mAP and 59.2\% NDS, reaching a level of accuracy comparable to the current state-of-the-art camera-based methods. In addition to accuracy, we also compare the overall computational efficiency of different camera-based detectors. For our model, we report two FPS measurements separated by “/”: the first corresponds to the complete pipeline including 2D detector, and the second measures only the 3D detection stage starting from 2D proposals. Notably, the 3D-only inference speed remains competitive despite the additional temporal stereo computation, reaching 19.4 / 13.8 FPS for the S-configuration with ResNet-50 and ResNet-101, and 11.3 / 8.4 FPS for the T-configuration. Although this represents a moderate slowdown compared with the MV2D baseline, the trade-off is acceptable given the substantial accuracy gains. It is also worth noting that we adopt a two-stage Faster R-CNN as the 2D detector, which introduces significant computational overhead due to its RPN, ROIAlign, and RCNN head, making it considerably slower than lightweight YOLO-style or DETR-style detectors. Thus, the overall FPS can be further improved by replacing the 2D detector with a more efficient alternative.

As illustrated in Tab. \uppercase\expandafter{\romannumeral2}, our method achieves up to 53.5\% mAP and 62.0\% NDS on the nuScenes test set, improving over the MV2D baseline by 2.4\% mAP and 2.6\% NDS, respectively. Compared with BEVStereo, a representative approach based on dense stereo matching, our method surpasses it by 1.0\% mAP and 1.2\% NDS, demonstrating that accurate depth perception can still be achieved even when stereo matching is performed only at the RoI level. On the AR2 validation set, as shown in Tab. \uppercase\expandafter{\romannumeral3}, our method not only surpasses the state-of-the-art sparse-query approaches such as the PETR series, but also outperforms two dense stereo–based models, BEVStereo and SOLOFusion, further validating the effectiveness of our framework.

\begin{table*}[t]
  \centering
  \caption{3D detection results on Argoverse 2 $val$ set. We evaluate 26 object categories with a range of 150 meters. Our model is marked in gray . Bold indicates the best performance.}
  \label{tab:3}
  \setlength{\tabcolsep}{4.6mm}{
  \begin{tabular}{c|cc|c|cc|ccccc}
  \toprule
  Method & Img Size & Backbone & Frames & mAP$\uparrow$ & CDS$\uparrow$ & mATE$\downarrow$ & mASE$\downarrow$ & mAOE$\downarrow$ \\
  \midrule
  PETR\cite{liu2022petr} & $960\times640$ & VoVNetV2 & 1 & 0.176 & 0.122 & 0.911 & 0.339 & 0.819 \\
  PETRv2\cite{liu2023petrv2} & $960\times640$ & VoVNetV2 & 2 & 0.188 & 0.129 & \textbf{0.801} & \textbf{0.314} & 0.778 \\
  BEVStereo\cite{li2023bevstereo} & $960\times640$ & VoVNetV2 & 2 & 0.146 & 0.104 & 0.847 & 0.397 & 0.901 \\
  SOLOFusion\cite{park2022time} & $960\times640$ & VoVNetV2 & 16+1 & 0.149 & 0.106 & 0.934 & 0.425 & 0.779 \\
  MV2D-T\cite{wang2023object} & $960\times640$ & VoVNetV2 & 2 & 0.189 & 0.132 & 0.830 & 0.346 & 0.740 \\
  \rowcolor{gray!30}StereoMV2D-T & $960\times640$ & VoVNetV2 & 2 & \textbf{0.194} & \textbf{0.138} & 0.811 & 0.325 & \textbf{0.723} \\ 
  \bottomrule
  \end{tabular} }
\end{table*}

\subsection{Ablation Study}
In this section, we design a series of ablation experiments to validate the effectiveness of each algorithmic component of our model. All experiments are conducted on the nuScenes validation set, trained for 24 epochs. Unless otherwise specified, the model adopts a ResNet-50 backbone with an input resolution of 1408×512 under the StereoMV2D-S configuration.

\textbf{Ablation on query generation strategies. } The primary innovation of this work lies in the way object queries are generated. Therefore, we first compare different query generation strategies and analyze their impact on detection accuracy. We evaluate the proposed mixed strategy against three alternatives: the fixed strategy, the mono (single-frame implicit) strategy, and the stereo (two-frame temporal stereo) strategy.
The mixed strategy follows PETR in initializing a fixed set of learnable queries, while the mono strategy adopts the Dynamic Object Query Generator (DOQG) from MV2D to produce initial reference positions from single-frame RoI features. The stereo strategy, in contrast, initializes queries solely using the reference positions generated by our sparse temporal-stereo query generator, without applying the confidence gating mechanism to fuse results from both branches.
Both MV2D and our method adopt a dynamic number of queries, determined by 2D detections. As shown in Tab. \uppercase\expandafter{\romannumeral4}, the fixed-query strategy improves as the number of queries increases, since a denser set of initial positions increases the likelihood of placing a query near an object. However, this improvement is limited and comes at substantial computational cost. Dynamic query generation strategies, guided by 2D detection priors, consistently exhibit superior performance. Compared to using only single-frame implicit depth inference or only two-frame stereo-based depth estimation, our mixed strategy achieves the best accuracy. This demonstrates that fusing the two complementary priors effectively mitigates their individual weaknesses—namely, depth ambiguity from single-frame prediction and the lack of reliable estimates for newly appearing or occluded objects when relying solely on stereo cues.

\begin{table}[t]
  \centering
  \caption{Comparison of different query generation strategies.}
  \label{tab:4}
  \setlength{\tabcolsep}{6.2mm}{
  \begin{tabular}{cccc}
  \toprule
  Strategy & Query Num & mAP$\uparrow$ & NDS$\uparrow$ \\
  \midrule
  fixed & 300 & 36.1 & 39.4 \\
  fixed & 900 & 36.2 & 39.9 \\
  fixed & 1500 & 36.6 & 40.6 \\
  mono & dynamic & 38.0 & 41.2 \\
  stereo & dynamic & 38.4 & 41.7 \\
  mixed & dynamic & \textbf{39.5} & \textbf{42.9}\\
  \bottomrule
  \end{tabular} }
\end{table}

\textbf{Design of dynamic confidence gating module. } To verify the effectiveness of the proposed dynamic confidence gating mechanism, we compare it against a simple baseline fusion strategy that directly averages the reference points produced by the two branches and a lightweight attention-based variant. The baseline (“Avg-Fusion”) computes the 3D reference prior as the arithmetic mean of the implicit and temporal-stereo priors for each RoI, without any reliability assessment. The attention variant (“Attn-Fusion”) replaces averaging with a single-head scaled dot-product attention over Top-K temporally matched RoIs: current RoI features are projected as queries, the Top-K historical features serve as keys/values, and the fused prior is obtained by the attention-weighted aggregation of the temporal branch’s candidates, followed by a linear rescore to produce the fusion weight. In contrast, our method (“Statistic-Gating”) estimates a per-RoI confidence for the temporal-stereo cue using low-order match statistics (e.g., max probability, entropy, dummy mass) combined with a compact appearance similarity to the Top-1 match, and then interpolates between the two priors with this confidence. Table \uppercase\expandafter{\romannumeral5} reports the performance comparison between the three fusion methods. Both Attn-Fusion and the proposed Statistic-Gating outperform the Avg-Fusion baseline, confirming the necessity of reliability-aware fusion. While the accuracy of Attn-Fusion and Statistic-Gating is close, Statistic-Gating attains slightly higher precision with notably lower computational overhead (no Q/K/V projections or Top-K attention maps), indicating that reliability can be captured effectively by simple distributional cues and a tiny MLP, yielding a better performance.

\begin{table}[t]
  \centering
  \caption{Comparison of three fusion
  methods.}
  \label{tab:5}
  \setlength{\tabcolsep}{9mm}{
  \begin{tabular}{ccc}
  \toprule
  Fusion Methods & mAP$\uparrow$ & NDS$\uparrow$ \\
  \midrule
  avg-fusion & 38.2 & 41.1 \\
  attn-fusion & 39.1 & 42.0\\
  statistic-gating & \textbf{39.5} & \textbf{42.9} \\
  \bottomrule
  \end{tabular} }
\end{table}

\textbf{Ablation on Motion-Aware Soft Matching. } To evaluate the effectiveness of the proposed motion-aware soft matching strategy, we compare it with two alternative correspondence modeling schemes. The simplest baseline performs appearance-only matching, where RoI features from the current and previous frames are directly compared based solely on visual similarity. A second variant augments this by incorporating 2D motion cues, using RoI coordinates on the image plane to estimate pseudo-motion and refine similarity computation, thus introducing limited geometric context. In contrast, our full MASM module leverages 3D motion–aware normalization to offer a geometrically grounded representation that facilitates more stable and discriminative correspondence estimation. The comparison among these three configurations allows us to assess how much 3D motion modeling contributes beyond appearance similarity and 2D positional heuristics. Tab. \uppercase\expandafter{\romannumeral6} demonstrates that MASM yields the most reliable associations, validating the benefit of incorporating aligned 3D priors into the matching process. 

\begin{table}[t]
  \centering
  \caption{Comparison of three RoI matching schemes.}
  \label{tab:6}
  \setlength{\tabcolsep}{8mm}{
  \begin{tabular}{ccc}
  \toprule
  Matching Methods & mAP$\uparrow$ & NDS$\uparrow$ \\
  \midrule
  appearance-only & 38.4 & 41.5 \\
  2D motion modeling & 38.7 & 41.9\\
  3D motion modeling & \textbf{39.5} & \textbf{42.9} \\
  \bottomrule
  \end{tabular} }
\end{table}

\textbf{Depth-matching execution region. } Since the purpose of performing stereo matching on an RoI pair is ultimately to obtain a single 3D reference point for that RoI, an engineering alternative is to avoid running stereo matching across the entire RoI region and instead estimate depth only at the RoI center, thereby reducing computational cost. To evaluate the impact of these two strategies, we compare their performance in terms of both accuracy and memory consumption. The results are summarized in Tab. \uppercase\expandafter{\romannumeral7}. Surprisingly, even when using only the RoI center for homography-based warping, the resulting point-level depth estimation is sufficient to achieve strong detection accuracy while reducing memory consumption. This makes it a viable alternative in scenarios where GPU memory is limited.

\begin{table}[t]
  \centering
  \caption{Comparison of different depth-matching execution regions.}
  \label{tab:7}
  \setlength{\tabcolsep}{4.8mm}{
  \begin{tabular}{ccccc}
  \toprule
  Execution Region & mAP$\uparrow$ & NDS$\uparrow$ & Memory \\
  \midrule
  RoI center & 38.8 & 42.0 & $\sim12000M$ \\
  entire RoI & \textbf{39.5} & \textbf{42.9} & $\sim15500M$ \\
  \bottomrule
  \end{tabular} }
\end{table}

\textbf{Ablation on 2D Detector. } To ensure a fair comparison with the baseline, we initially adopt the same 2D detector as MV2D, i.e., Faster R-CNN. However, as discussed in Section \uppercase\expandafter{\romannumeral4}-C, Faster R-CNN is a two-stage detector and is inherently less efficient than one-stage detectors in terms of inference speed. Therefore, we conduct an ablation study on the choice of the 2D detector by replacing Faster R-CNN with the one-stage detector YOLOX. In addition, to explore the upper bound of StereoMV2D’s inference efficiency, we further reduce the input image resolution. The corresponding experimental results are reported in Table \uppercase\expandafter{\romannumeral8}. As shown, substituting Faster R-CNN with the lightweight YOLOX detector significantly improves the overall inference speed, increasing from 3.9 FPS to 8.2 FPS, while still maintaining competitive detection accuracy. When the input image resolution is further reduced to $704\times256$, the detection accuracy decreases due to the loss of high-resolution visual details; nevertheless, the degradation remains manageable, and the inference speed reaches up to 27.6 FPS. These results demonstrate that StereoMV2D can flexibly adapt to resource-constrained scenarios by replacing the 2D detector and adjusting the input resolution.

\begin{table}[t]
  \centering
  \caption{Comparison of different 2D detectors.}
  \label{tab:8}
  \setlength{\tabcolsep}{2mm}{
  \begin{tabular}{cccccc}
  \toprule
  Resolution & backbone & detector & mAP$\uparrow$ & NDS$\uparrow$ & FPS$\uparrow$ \\
  \midrule
  $704\times256$ & ResNet-50 & YOLOX & 37.6 & 40.7 & 27.6 \\
  $1600\times640$ & ResNet-50 & YOLOX & 39.2 & 42.8 & 8.2 \\
  $1600\times640$ & ResNet-50 & Faster-RCNN & 39.5 & 42.9 & 3.9 \\
  \bottomrule
  \end{tabular} }
\end{table}

\subsection{Qualitative Analysis}
\begin{figure}[!t]
\centering
\includegraphics[height=0.24\textheight]{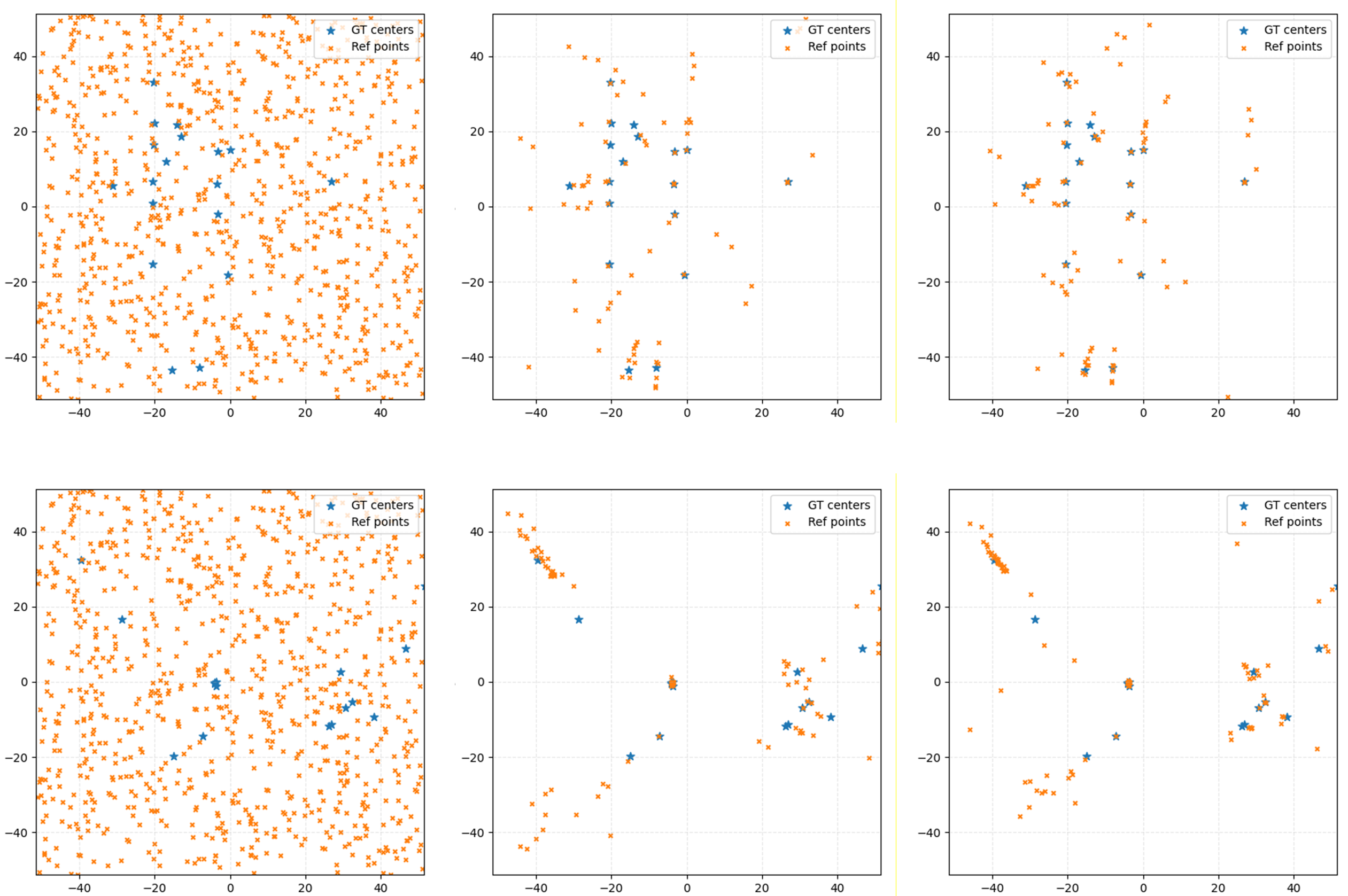}
\caption{Visualization of query locations generated by different query initialization methods. The first column corresponds to the fixed-number learnable query initialization, the second column shows the single-frame implicit query generator from MV2D, and the third column presents our proposed sparse temporal-stereo–based query generation method.}
\label{fig_3}
\end{figure}

\begin{figure*}[!t]
\centering
\includegraphics[width=7.1in]{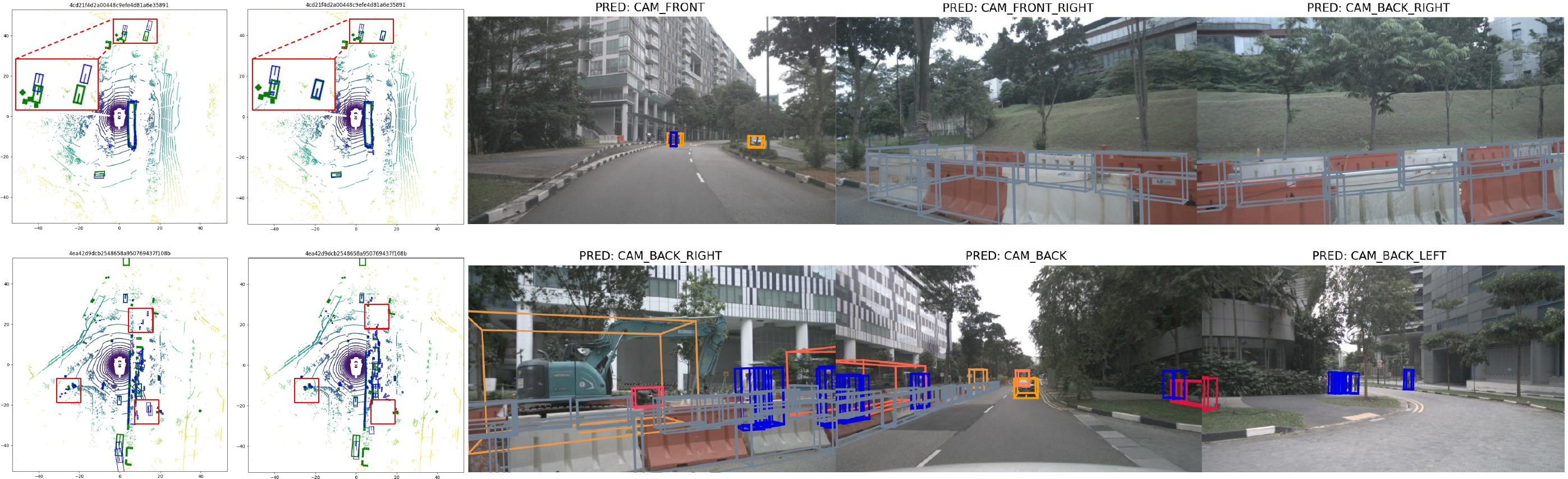}
\caption{Visualization of 3D detection results. On the left, we compare the BEV predictions of our method with those of a representative dense-BEV 3D detection baseline. Green boxes denote ground truth, and blue boxes denote predictions. The first column shows the baseline results, while the second column presents StereoMV2D’s outputs. Regions highlighted with red rectangles indicate areas with noticeably improved localization or reduced false positives. On the right, we visualize the 3D detection results of StereoMV2D across multiple camera views, where bounding boxes of different colors correspond to different object categories.}
\label{fig_4}
\end{figure*}

In this section, we provide qualitative results on the nuScenes validation set to illustrate the improvements brought by our method. In Fig. 3, we compare the query locations generated by different query initialization strategies. Each row corresponds to the same scene, while each column represents a different query generation method. Compared with the commonly used learnable query initialization in sparse query–based detectors, methods that leverage 2D detections to provide positional priors clearly demonstrate a superior ability to cover regions around true object locations, even with significantly fewer queries, enabling more efficient and focused feature learning. Relative to the baseline, our query generation approach produces query positions that cluster more tightly around actual objects, indicating that the proposed sparse temporal stereo mechanism successfully enhances depth awareness and yields more accurate spatial priors.

Fig. 4 further illustrates 3D object detection results from StereoMV2D. From the multi-view images, we observe that our method can accurately detect different object categories even in challenging scenarios such as dense traffic, partial occlusion, and long-range objects. The BEV visualizations on the left show that StereoMV2D produces fewer false positives and exhibits reduced localization bias, particularly for small or distant objects. This improvement stems from two key factors: (1) the strong foreground filtering capability of the 2D detector, and (2) the precise object localization enabled by our temporal stereo–driven depth reasoning.

\section{Conclusion}
In this paper, we presented StereoMV2D, a temporal stereo–driven sparse query framework for multi-view 3D object detection. By integrating object-centric RoI-level temporal stereo into a query-based paradigm, our method effectively alleviates the depth ambiguity inherent to single-frame detection while preserving the efficiency advantages of sparse feature aggregation. The proposed Motion-Aware Soft Matching and RoI-Level Temporal Stereo Matching modules jointly enable precise temporal correspondence and lightweight stereo reasoning, and the dynamic confidence gating mechanism adaptively balances monocular and stereo priors to ensure robust performance under occlusion and object emergence. Extensive experiments demonstrate that StereoMV2D achieves good accuracy–efficiency trade-offs and consistently surpasses the baseline MV2D across multiple configurations.

Despite its advantages, StereoMV2D still relies on a 2D detector, which limits end-to-end efficiency and contributes significantly to the overall inference latency. Moreover, although RoI-level stereo greatly reduces computational cost compared to dense stereo matching, it does not fully exploit long-range temporal information that may further improve depth stability. In future work, we plan to extend our temporal modeling to multi-frame geometric reasoning to learn more flexible motion priors. Another promising direction is to integrate additional sensing modalities such as 4D radar or LiDAR to achieve more robust multimodal detection.

\vfill

\end{document}